\documentclass[
]{ceurart}
\pdfoutput=1
\usepackage{listings}
\usepackage{fancyvrb}
\usepackage{paralist}
\usepackage{soul}
\usepackage{todonotes}
\begin{document}

\copyrightyear{2023}
\copyrightclause{Copyright for this paper by its authors.
  Use permitted under Creative Commons License Attribution 4.0
  International (CC BY 4.0).}

\conference{Companion Proceedings of the 16th IFIP WG 8.1 Working Conference on the Practice of Enterprise Modeling and the 13th Enterprise Design and Engineering Working Conference, November 28 – December 1, 2023, Vienna, Austria}

\title{A Digital Twin prototype for traffic sign recognition of a learning-enabled autonomous vehicle}

%\tnotetext[1]{You can use this document as the template for preparing your publication. We recommend using the latest version of the ceurart style.}

\author[1]{Mohamed AbdElSalam}[%
email=mohamed.abdelsalam@siemens.com,
]
\address[1]{Siemens EDA,     Cairo, Egypt}
\author[1]{Loai Ali}[%
email=loai.ali@siemens.com,
]
\cormark[1]
\author[2]{Saddek Bensalem}[%
email=saddek.bensalem@univ-grenoble-alpes.fr,
]
\address[2]{ Universit\'{e} Grenoble Alpes, VERIMAG,
     Grenoble, France}
\author[2]{Weicheng He}[%
email=weicheng.he@univ-grenoble-alpes.fr,
]
%\address[2]{}{Universit\'{e} Grenoble Alpes, VERIMAG,
     %Grenoble, France}
\author[3]{Panagiotis Katsaros}[%
email=katsaros@csd.auth.gr,
]
\address[3]{ Aristotle University of Thessaloniki, Thessaloniki, Greece\\
}
\author[3]{Nikolaos Kekatos}[%
email=nkekatos@csd.auth.gr,
]
%\address[4]{ DRAXIS Environmental S.A., Thessaloniki, Greece\\}
\author[4]{Doron Peled}[%
email=doron.peled@gmail.com,
]
\address[4]{     Bar Ilan University,  Ramat Gan, Israel
}
\author[3]{Anastasios Temperekidis}[%
email=anastemp@csd.auth.gr,
]
\author[2]{Changshun Wu}[%
email=changshun.wu@univ-grenoble-alpes.fr,
]
\cortext[1]{Corresponding author.}
%\fntext[1]{These authors contributed equally.}

\begin{abstract}
 In this paper, we present a novel digital twin  prototype for a learning-enabled self-driving vehicle. The primary objective of this  digital twin   is to perform traffic sign recognition and lane keeping. The digital twin architecture relies on co-simulation and uses the Functional Mock-up Interface and SystemC Transaction Level Modeling  standards. The  digital twin  consists of four clients, i) a vehicle model that is designed in Amesim tool, ii) an environment model developed in Prescan, iii) a lane-keeping controller designed in Robot Operating System, and iv) a perception and speed control module developed in the formal modeling language of BIP (Behavior, Interaction, Priority). These clients interface with the digital twin platform, PAVE360-Veloce System Interconnect (PAVE360-VSI). PAVE360-VSI acts as the co-simulation orchestrator and is responsible for synchronization, interconnection, and data exchange through a server. The server establishes connections among the different clients and  also ensures adherence to the  Ethernet protocol. We conclude with illustrative   digital twin   simulations and recommendations for future work.
 % The communication between the server and both remote clients is established using the EMU TLM API raw C client portal, which is used to establish TLM-2.0 channels with the FabricServer TLM server.
 %Ethernet frames are used that are connected to a network interconnect,    
\end{abstract}


\maketitle

\section{Introduction}

\emph{Digital Twins} (DTs)~\cite{grieves2016origins}  have gained significant attention in both academia and industry.  Typically defined as a virtual representation of a physical asset, process, and system, a DT  serves diverse purposes~\cite{grieves2019virtually}. DT applications  have been found in various domains, like product lifecycle management and manufacturing~\cite{Grieves15}. The main objective of a DT is to generate virtual/digital models of physical objects  to simulate their behaviors~\cite{grieves2019virtually}. Conventionally, DTs involve three elements:
i) a physical entity, i.e. an artifact like a vehicle, a product, a process, or a system in real
space, ii) a virtual entity, i.e. a virtual  representation of
the physical entity, and iii) a channel that links these two entities  and can transfer information from one to the other. 

The digital twin concept enables creating and refining a virtual counterpart before the physical one exists~\cite{grieves2017digital}. If the digital system meets certain criteria, the physical product can be manufactured and linked to its DT using sensors, often referred to as a digital twin prototype (DTP)~\cite{grieves2023digital}. The DTP includes all  models and processes for realizing the physical entity~\cite{jones2020characterising}.

Digital twins are increasingly applied in the automotive domain~\cite{allamaa2022sim2real,piromalis2022digital, esen2023simulation,tottrup2022using}, with ongoing development in new DT technologies and frameworks~\cite{ chaudhuri2023predictive,hartmann2022executable,torzoni2024digital,frasheri2023addressing,deakin2023smart}. In this work, we present our design of a digital twin of a learning-enabled autonomous vehicle. The motivation for this DT stems from the FOCETA project, with its  high-level schematic and intended functionality  outlined   in~\cite{10.1007/978-3-031-46002-9_15}. Our specific contributions in this paper involve:
\begin{compactitem}
\item developing a complete digital twin prototype of a learning-enabled autonomous vehicle that seamlessly integrates and interconnects multiple components and subsystems. We  leverage the PAVE360-VSI digital twin platform and Ethernet protocol.
\item designing a 15 Degrees of Freedom (15DoF)  vehicle model in Amesim. The model is exported as a Functional Mock-up Unit (FMU) for interoperability purposes.
\item generating an environment model in Prescan that specifies the road path, the sensors, and the driving scenario.
\item creating a perception module, based on the YOLOX algorithm, which can detect and classify traffic signs.
\item implementing a steering controller in the Robot Operating System (ROS) for lane keeping.
\end{compactitem}
The rest of the paper is structured as follows. In Section~\ref{sec:languages_tools}, we introduce
the key technologies, standards, and tools employed  in the co-simulation and
digital twin architecture. In Section~\ref{sec:dt}, we present all the DT components and building
blocks designed to execute DT simulations. The paper concludes in
Section~\ref{sec:conclusion}.

\section{Key enabling technologies for Digital Twins}\label{sec:languages_tools}

%\subsection{Co-simulation} 
\paragraph{Co-simulation.} Co-simulation is a practical technique for simulating heterogeneous systems. It permits the modeling and simulation of different components of a system using various tools and methods. The global simulation of a coupled system can then be achieved by composing the simulations of its parts. Co-simulation also allows  the joint simulation of loosely coupled stand-alone sub-simulators. To guarantee that the submodels and sub-simulators can work seamlessly together,  standardized interfaces are commonly used~\cite{gomes2017co,gomes2018co,talasila2023digital}.

%\subsection{Functional Mock-up Interface (FMI)}
\paragraph{Functional Mock-up Interface (FMI).} A widely used standard  for co-simulation is the FMI standard~\cite{blockwitz2012functional,junghanns2021functional}. The FMI standard defines an  Application Programming Interface (API) and the model elements referred to as Functional Mock-up Units (FMUs), which adhere to this API. Each FMU can be seen as a black box that implements the methods defined in the FMI API. 
To run a group of interconnected FMUs, an orchestrator, also called a master algorithm or FMI Master, is required. Its purpose is to manage and synchronize their execution.

%\subsection{Transaction Level Modeling (TLM)} 
\paragraph{Transaction Level Modeling (TLM).} 
The SystemC-TLM 2.0 standard simplifies communication and computation in industrial simulation models by using channels. These channels abstract unnecessary details, resulting in faster simulations and facilitating high-level design validation. Transactions are initiated through the functions in the channel model interface.

The primary focus of the TLM transaction API is the versatile \emph{generic payload} transaction, which can be adapted to various application domains~\cite{frank2005transaction}. TLM is suitable for a wide range of domains, including digital and analog simulations, as well as digital communication protocols like CAN, Ethernet, AXI, and PCIe.

%\subsection{Behavior, Interaction, Priority (BIP)}
\paragraph{Behavior, Interaction, Priority (BIP).} 
BIP is a  modeling framework for rigorous system design~\cite{basu2011rigorous}. In BIP, complex systems are constructed by coordinating the behavior of atomic components. BIP \textit{atomic components} are transition systems  with ports and  variables.  An atomic component may contain control locations, variables, communication ports, and transitions.  Composite components are built by composing multiple atomic components. Interactions between components are defined by connectors, which specify sets of interactions.

During the execution of a BIP interaction, all components that participate in the interaction, i.e., have an associated port that is part of the interaction, must execute their corresponding transitions simultaneously. All components that do not participate in the interaction, do not execute any transition and thus remain in the same control location.

%\subsection{Digital twin platform}
\paragraph{Digital Twin platform.}

PAVE360-Veloce System Interconnect (PAVE360-VSI)~\cite{veloce} is a digital twin platform  with network interconnect and hardware emulation capabilities for autonomous systems. It supports diverse client connections, enabling protocol-agnostic and protocol-aware links between mixed-fidelity models. PAVE360-VSI is suitable for Model-in-the-Loop (MiL), Software-in-the-Loop (SiL), and Hardware-in-the-Loop (HiL) verification, with a focus on pre-silicon verification~\cite{abdelsalam2019verification}.

PAVE360-VSI provides a co-simulation architecture for DTs, which complies with Functional Mock-up Interface (FMI) and Transaction-Level Modeling (TLM) standards. Simulations advance in discrete time steps, leveraging a server/client structure within the PAVE360-VSI architecture. This design ensures interoperability among DT components, synchronizes network communication and facilitates data transfer using multiple protocols. 

The core assumption is that any external simulator and foreign model can be integrated with a third-party API customized for the application's needs. This API is  linked to a TLM fabric portal interface known as the "Gateway"~\cite{uvm}. The Gateway acts as an entry point to the DT backplane, enabling transactional communication. Each external simulator or foreign model is  a client process connected to the shared DT backplane. This backplane acts as the timekeeper, overseeing all time advancement operations to maintain synchronization among client processes and ensure deterministic behavior.

PAVE360-VSI  has been extended to facilitate formal modeling. A "BIP Gateway" has been created in~\cite{temperekidis2022towards}. Runtime verification can be used within the architecture via FMI~\cite{temperekidis2022runtime}. 

\section{Developing a Digital Twin for traffic sign recognition and lane keeping}\label{sec:dt}

\begin{figure}[htb!]
    \centering
    \includegraphics[width=0.8\textwidth]{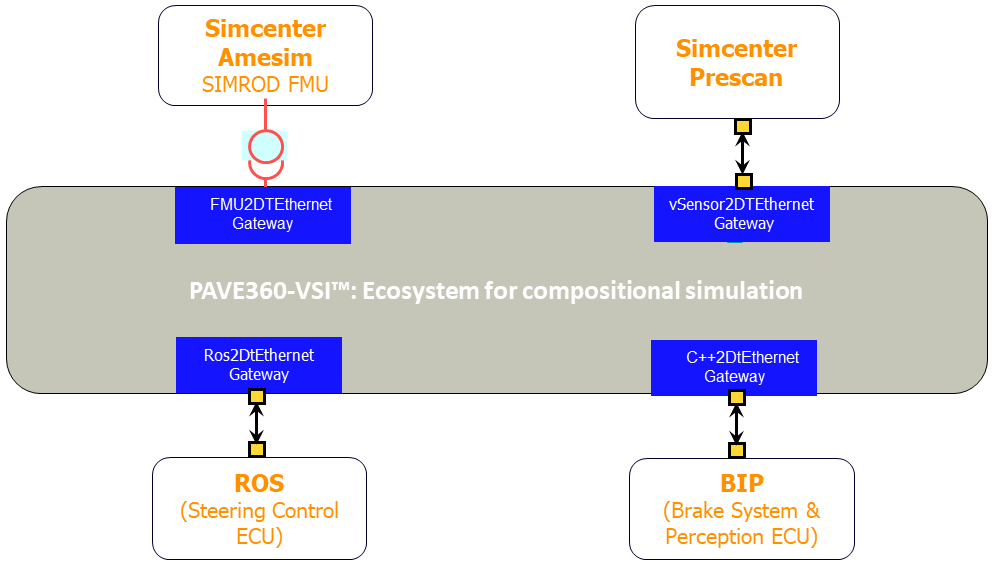}
    \caption{Digital twin prototype for traffic sign recognition of a learning-enabled vehicle. Architecture: PAVE360-VSI DT platform containing a server and four clients/components; a vehicle model modeled in AMESIM, a control ROS subsystem,
a control module modeled in BIP, and a Prescan model. All tools are
co-simulated and interconnected via gateways.}
    \label{fig:arch-dt}
\end{figure}

The digital twin comprises four  heterogeneous components that communicate through the PAVE360-VSI platform. Figure~\ref{fig:arch-dt} presents the overall DT prototype of our learning-enabled vehicle with lane-keeping and traffic sign recognition capabilities. Each component has a different role, i.e., vehicle modeling,  environment  modeling, lane keeping, traffic sign recognition and speed regulation, each  detailed in separate sections.

Data exchange among these components occurs via the Ethernet protocol. Gateways play a crucial role in converting data into simulated Ethernet frames.
These exchanged frames adhere to the structure of a typical (physical) Ethernet network connection, incorporating familiar fields such as MAC headers, payloads, and checksums.
The payload of the exchanged Ethernet frames consists of a serialized bitstream representing the data to be exchanged between the communicating entities. This payload is designed to accommodate various data types, such as floats, integers, strings, and other relevant data structures. This flexibility enables the digital twin to exchange a diverse range of information, facilitating a comprehensive simulation of the ego vehicle's behavior and its interactions with the environment.

\subsection{Vehicle model: Amesim}
The vehicle is described by a high-fidelity dynamical model (with 15 degrees of freedom) in Amesim. It is adapted from the Simrod SIEMENS model~\cite{amesim}. It is encapsulated as an FMU to connect with the interconnect and we use an FMU to ethernet gateway.

The FMU to ethernet gateway has a dual role and it supports the bi-directional transfer of data from the FMU to the interconnect and vice versa. In this case, its primary task is to receive an ethernet frame that contains information about the ego vehicle’s coordinates, brake, and throttle commands. The gateway decapsulates this information from the ethernet frame (keeping only the payload) and feeds it into the FMU model. The payload consists of three float variables representing the ego vehicle’s coordinates, brake, and throttle commands. This model maintains a representation of the ego-vehicle dynamics. Within the FMU, we execute a step, advancing the simulation by a fixed time step. 

\begin{figure}[tb!]
    \centering
    \includegraphics[width=0.68\textwidth]{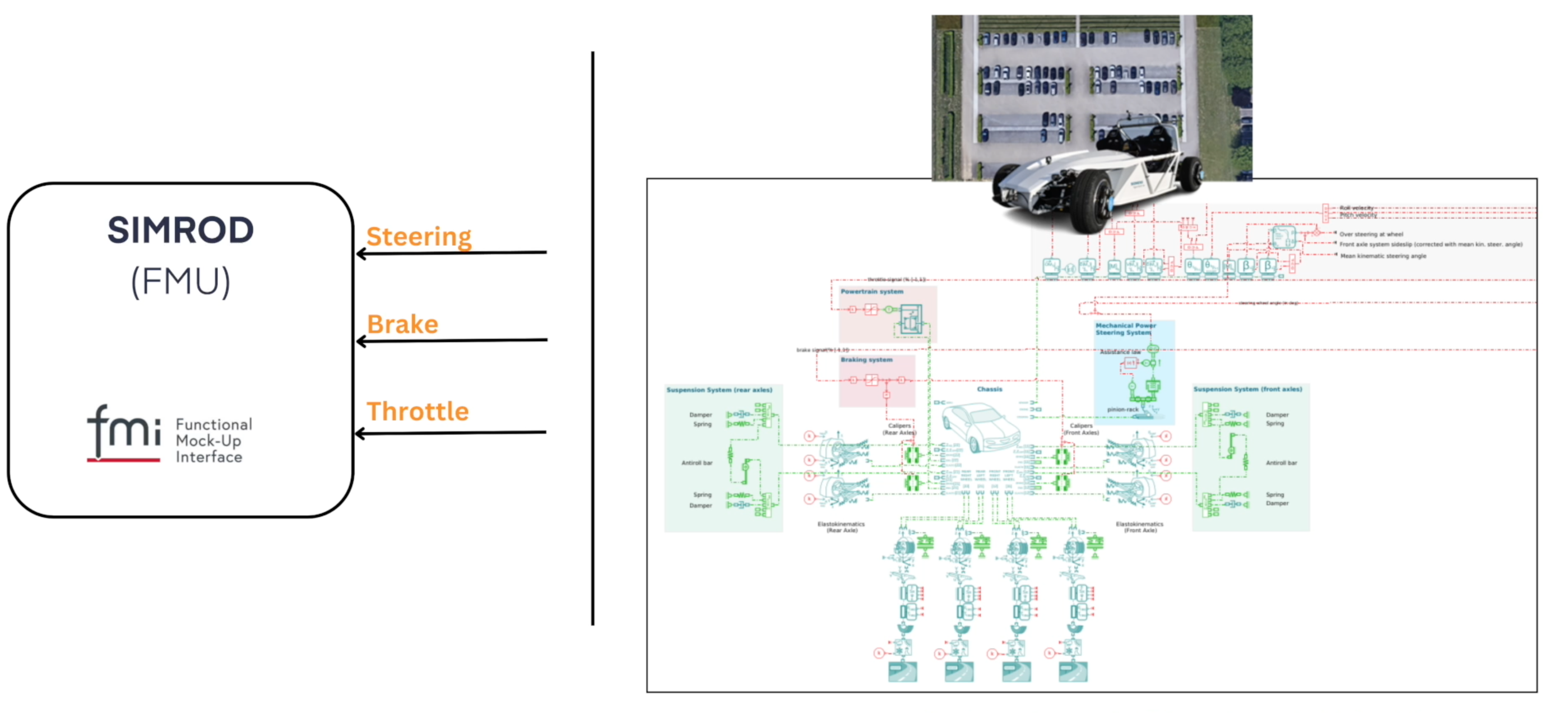}
    \caption{Vehicle Model in Amesim; the left part shows the inputs and outputs of the FMU client, while the right part displays a visual representation of the Simrod Amesim FMU.}
    \label{fig:FMU_label}
\end{figure}

Figure \ref{fig:FMU_label} provides some  details of the Amesim model and consists of two parts. The left part shows a visual representation at a top level for the FMU client representing its inputs/outputs, while the right part shows the Simrod Amesim FMU.
\subsection{Steering control: ROS}

The Robot Operating System (ROS) client plays a key role in the navigation of the ego vehicle. It receives real-time coordinates of the ego vehicle’s position, providing a continuous update of its location within the Prescan environment.
The ROS gateway works as follows. It receives an ethernet frame that encapsulates the current coordinates of the ego-vehicle. The payload of the received ethernet frame is a stream of bytes for three float variables representing the ego vehicle's coordinates, the ego vehicle's speed, and acceleration. The ROS gateway then decodes the data from the frame and sends this information to other ROS nodes. These nodes are designed with a specific algorithm for steering the car. They process the positional data and convert it into actionable steering commands.
These steering commands are subsequently published by the ROS nodes. The ROS gateway receives these commands, encapsulates them within an Ethernet frame, and transmits this information. The payload of the transmitted Ethernet frame contains a float variable representing the new steering value of the ego vehicle.
This ensures that the ego vehicle navigates smoothly and accurately through the virtual world, responding appropriately to the changing conditions and scenarios.

\begin{figure}[tb!]
    \centering
    \includegraphics[width=0.75
    \textwidth]{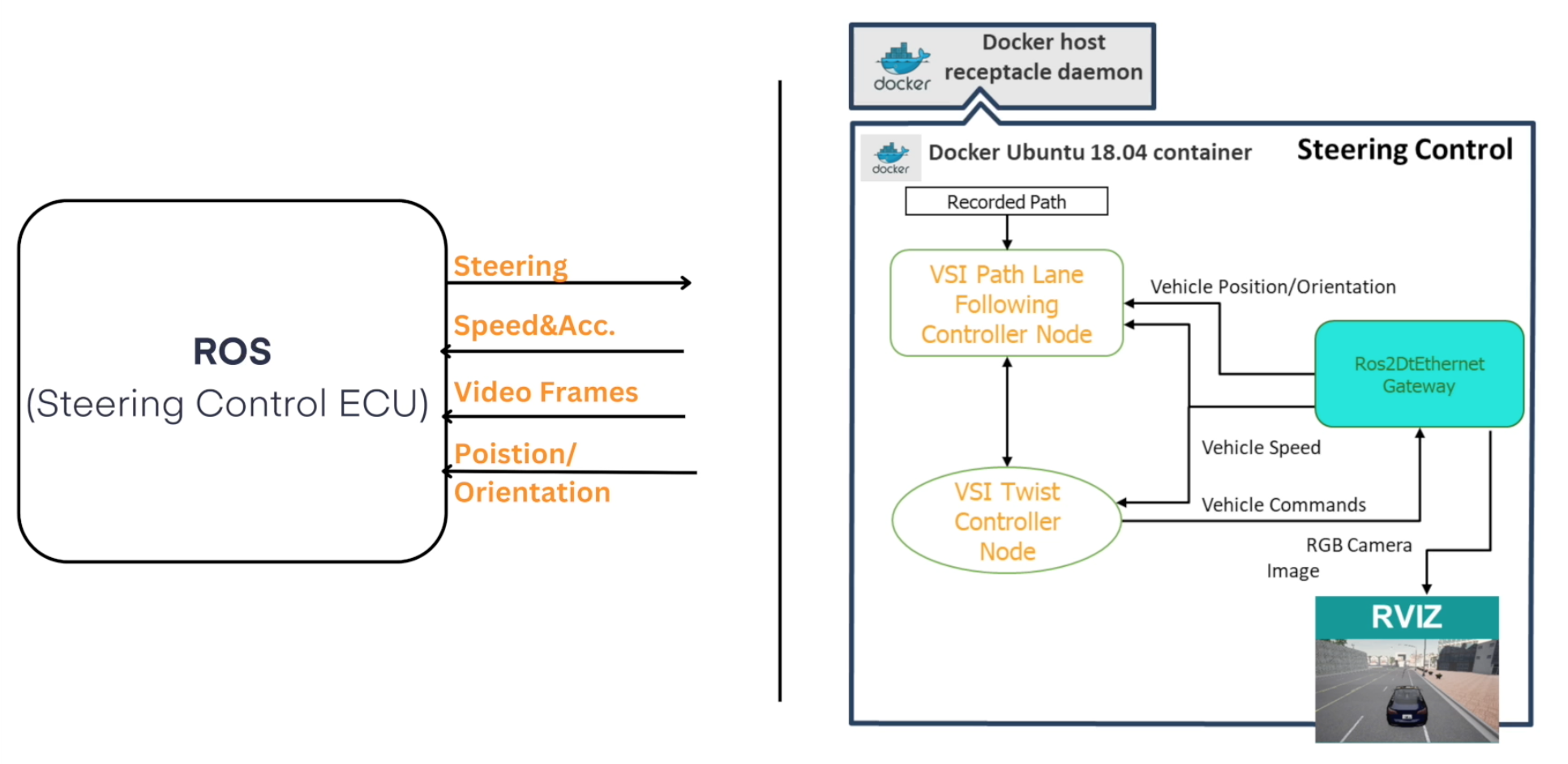}
    \caption{Digital Twin component - ROS client - steering control module.}
    \label{fig:ros-label}
\end{figure}

Figure \ref{fig:ros-label} shows how ROS subsystem is structured and containerized in Docker.  The left part shows a visual representation at a top level for the ROS client representing its inputs/outputs, while the right part shows the ROS subsystem.

\subsection{Environment model: Prescan}

The Prescan client serves as a digital representation of a real-world environment, designed to mirror actual driving conditions. This virtual world includes roads, traffic signs, and other elements typically encountered during a drive.

The main actor of this Prescan scenario is the ego vehicle, a digital representation of a real-world vehicle. This vehicle is equipped with an advanced RGB camera sensor, which captures video frames of the surrounding environment. These frames provide visual data that is disseminated to various clients within the network. This continuous stream of information forms the basis for decision-making processes in other clients, enabling them to react and adapt in real time to the changing conditions within the Prescan environment.

The vSensor (virtual sensor) to Ethernet gateway, implemented as an S-function within the Prescan client, plays a key role in this process. It encapsulates the video frame captured by the RGB camera sensor mounted on top of the ego-vehicle inside an Ethernet frame and transmits it. The video frames that are sent in the Ethernet frame are represented by a stream of bytes representing the RGB values of the pixels of the image that is captured by the RGB camera sensor. In addition, the gateway sends another Ethernet frame containing the current coordinates and speed of the ego-vehicle, conveyed as float variables in the  payload. Finally, the gateway  receives two Ethernet frames containing the steering angle, brake,  and throttle values. 

\begin{figure}[htb!]
    \centering
    \includegraphics[width=0.75\textwidth]{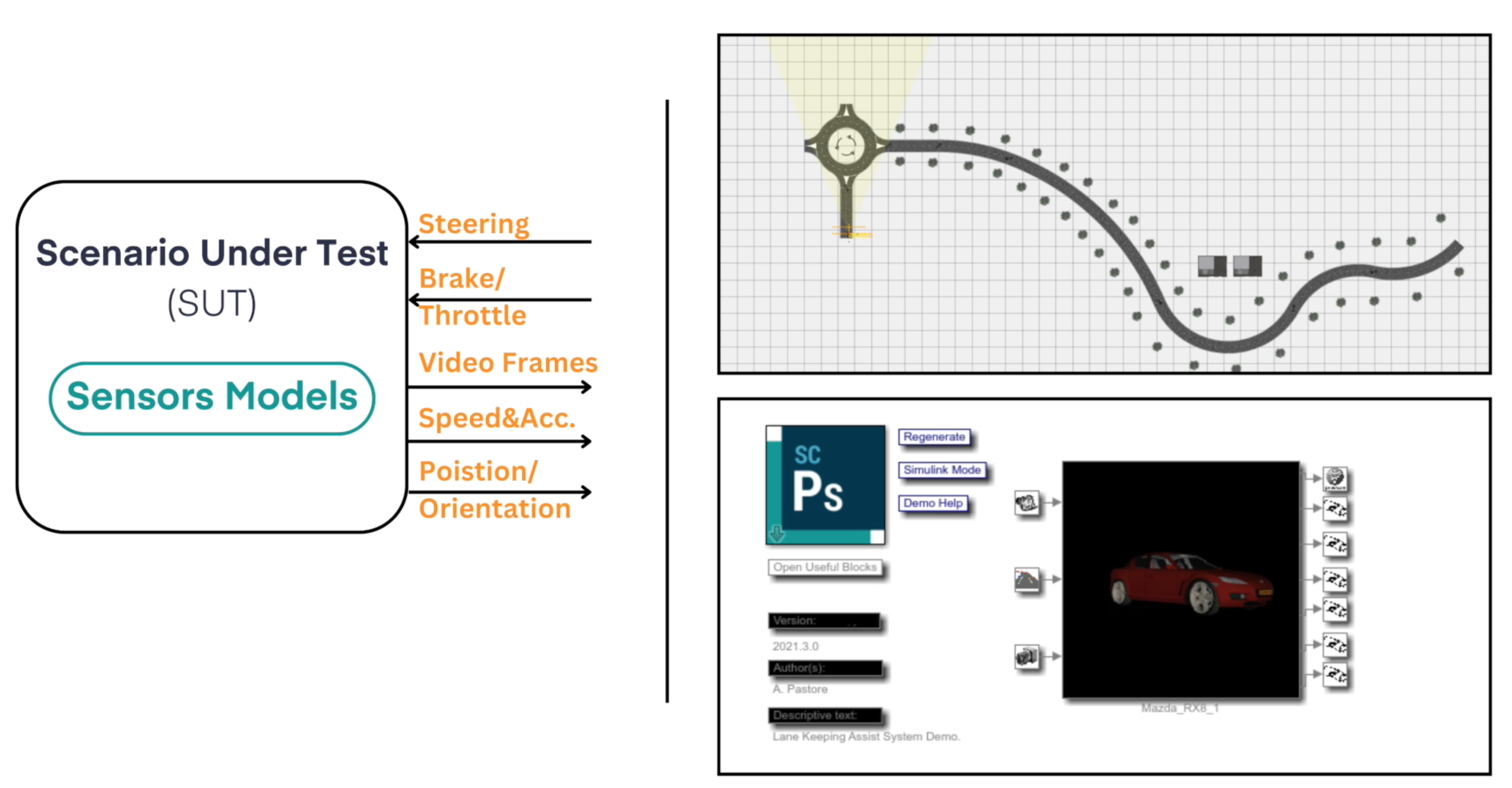}
    \caption{Digital Twin component - Prescan scenario - environment model and driving track.}
    \label{fig:prescan-label}
\end{figure}

In Figure \ref{fig:prescan-label}, we show the Prescan scenario under study. The left part shows a visual representation at a top level for the prescan client representing its inputs/outputs, while the right part shows the Prescan experiment setup and also the ego-vehicle used.
\subsection{Perception and speed control module: BIP}
The BIP client, another DT component, serves two essential functions  for the ego vehicle. Firstly, it employs a YOLOX deep learning model~\cite{ge2021yolox} to process video frames received from the ego vehicle's RGB camera sensor. These frames, transmitted as an Ethernet frame from the Prescan client, pass through the C\texttt{++} to Ethernet gateway within the BIP client.  The video frame within the Ethernet frame is represented as a byte sequence capturing RGB values of the pixels. The gateway decapsulates this data and forwards it to YOLOX. The YOLOX model has been trained to detect speed limits and identify Prescan traffic signs. %The video frame inside the received Ethernet frame is represented as a sequence of bytes representing the RGB values of the pixels of video frame. The gateway decapsulates this information and feeds it to YOLOX. This model is trained to detect speed limits from these frames, identifying the Prescan traffic signs.

The second function of the BIP client is to regulate the velocity of the ego vehicle  based on the  detected speed limit and the current speed of the ego vehicle. The current speed of the ego vehicle is a float variable received via  the C\texttt{++} to Ethernet gateway. A Proportional-Integral (PI) controller then uses both pieces of information to apply either brake or throttle commands based on the difference between the detected speed limit and the vehicle’s current speed.

\begin{figure}[tb!]
    \centering
    \includegraphics[width=0.8\textwidth]{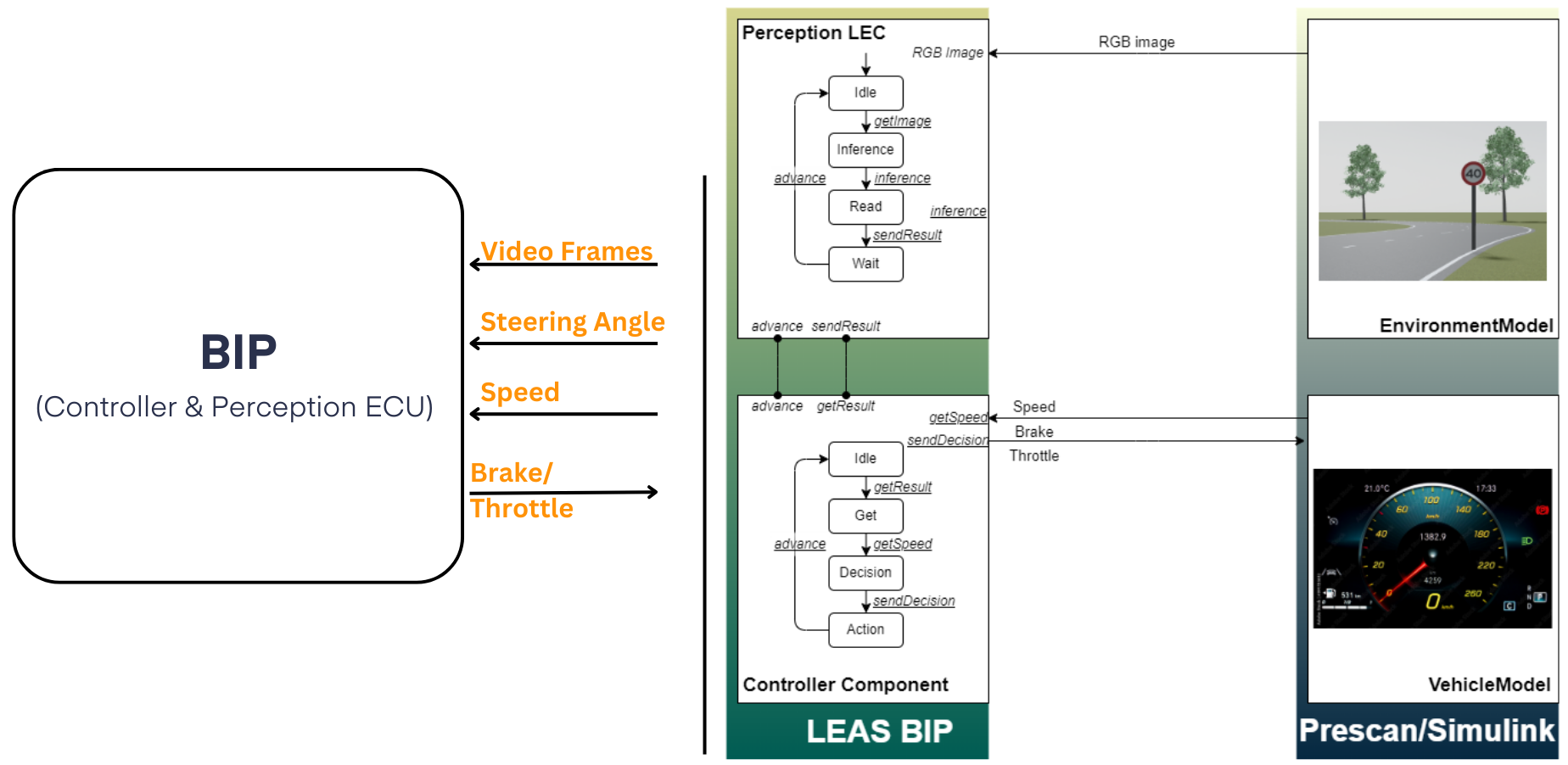}
   \caption{Digital Twin component - BIP module - speed control regulation and perception with YOLOX.}
    \label{fig:bip-label}
\end{figure}

Figure \ref{fig:bip-label} provides a visual representation of the key functionalities of the BIP client. The left part illustrates the client's inputs/outputs, while the right part outlines the internal interactions within the BIP client.
\subsection{Orchestration: PAVE360-VSI platform}
The PAVE360-VSI platform serves as the cornerstone for co-simulation and digital twin orchestration, offering functions that seamlessly integrate all clients. %It is responsible for managing and routing the data exchanged between different clients, ensuring efficient and effective communication within the network.

\begin{figure}[b!]
    \centering
    \includegraphics[width=0.99\textwidth]{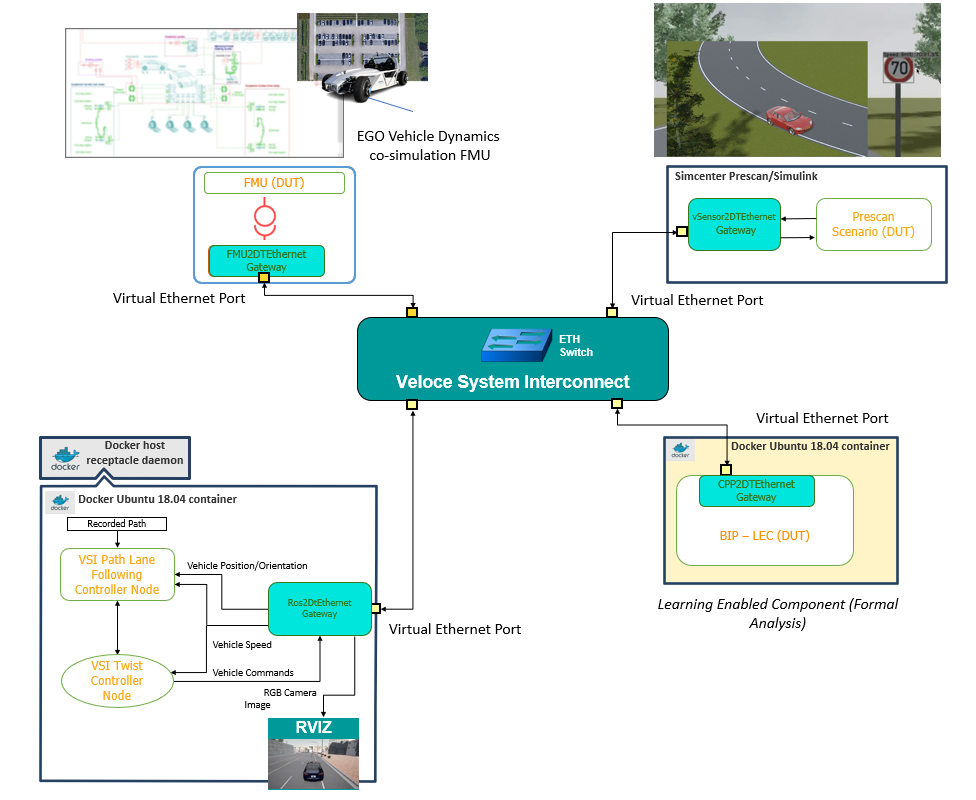}
        \caption{Architecture of the digital twin showing all clients' connections.}
    \label{fig:connections}
\end{figure}

\begin{figure}[tb!]
    \centering
    \includegraphics[width=0.95\textwidth]{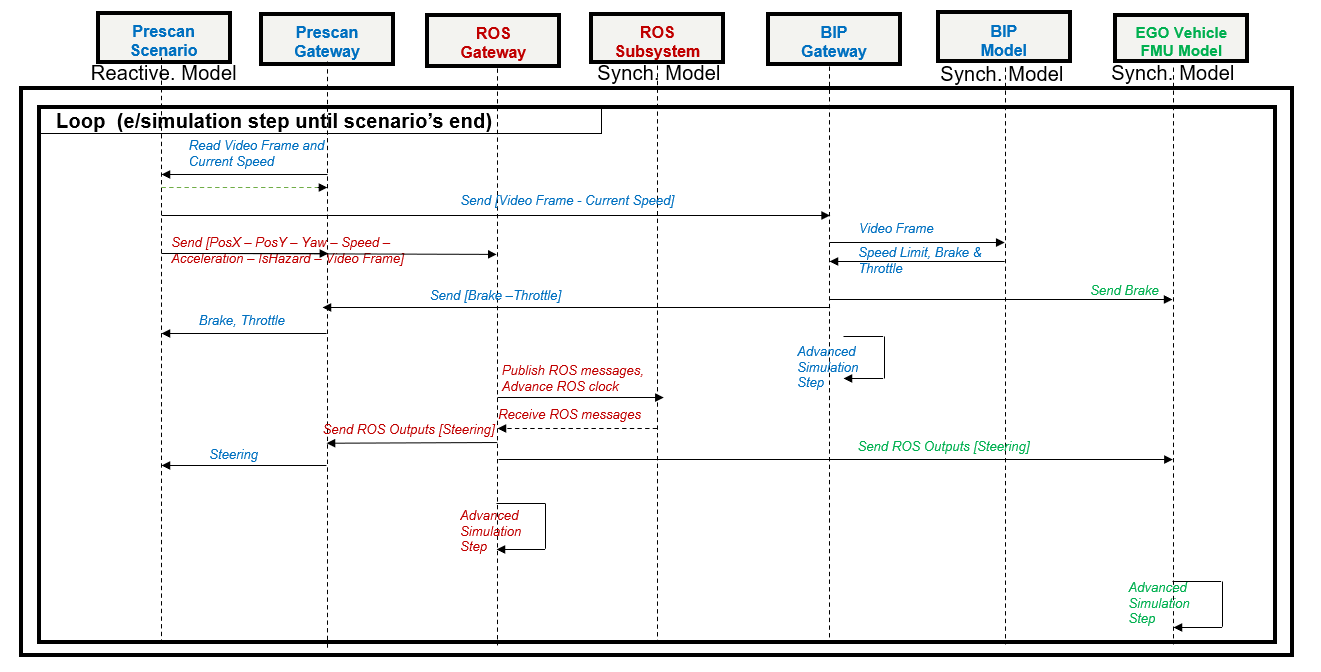}
  \caption{Message sequence diagram of the digital twin showing the order of the component calls.}
 \label{fig:sequence}
 \end{figure}

In this work, to realize the digital twin prototype, we employed a simulated Ethernet switch within the PAVE360-VSI digital twin platform for data exchange. All four clients connect to this switch through gateways, responsible for encapsulating and decapsulating data into/from Ethernet frames. The orchestrator's switch then takes control, routing data to the appropriate client (see  Figure~\ref{fig:connections}). The orchestrator is also responsible for synchronizing all clients based on a given message sequence diagram. It ensures that all components operate on the same simulated time, maintaining safe and stable operation and coordination within the system. Figure~\ref{fig:sequence} depicts the message  state sequence we designed and implemented.

\subsection{Digital Twin simulation}
\begin{figure}[tb!]
    \centering
    \includegraphics[width=0.95\textwidth]{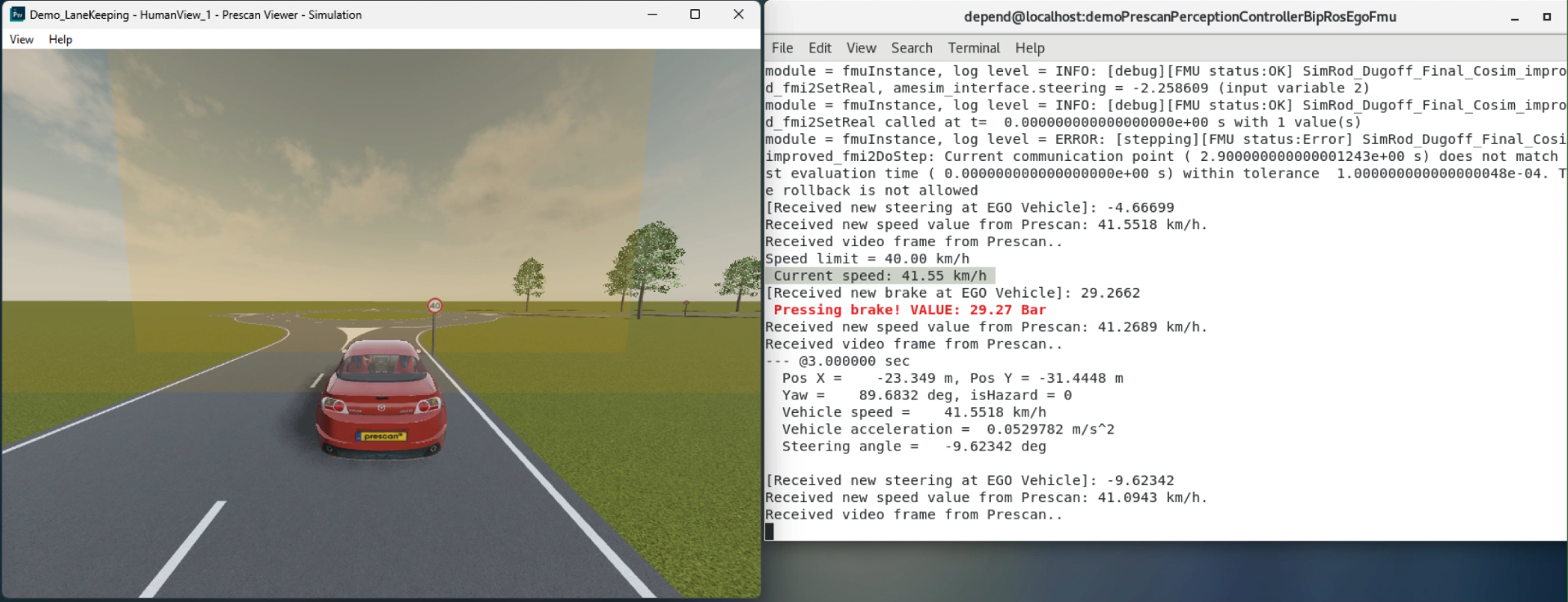}
    \caption{Digital twin simulation of a learning-enabled vehicle  showcasing a successful  detection of a \emph{40 km/h} speed limit sign.}
    \label{fig:rec-1-label}
\end{figure}

\begin{figure}[htb!]
    \centering
    \includegraphics[width=0.95\textwidth]{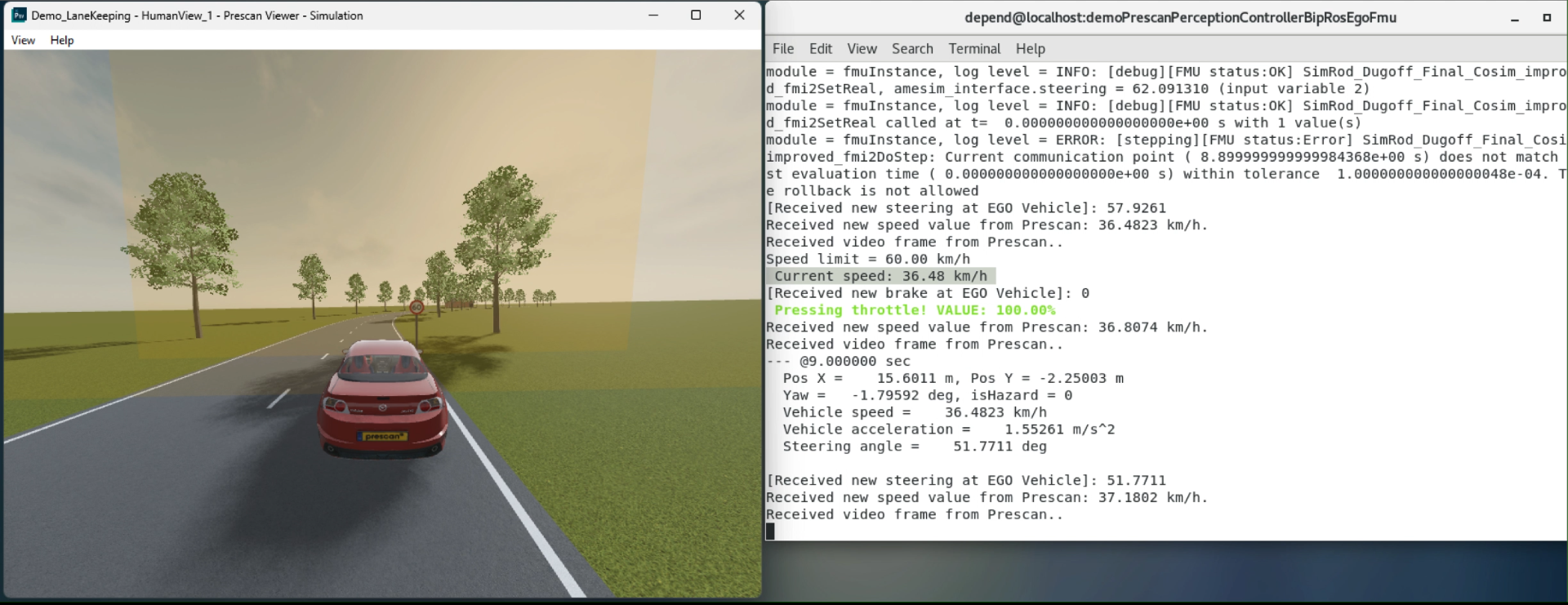}
    \caption{Digital twin simulation of a learning-enabled vehicle  showcasing a successful  detection of a \emph{60 km/h} speed limit sign.}
    \label{fig:rec-2-label}
\end{figure}

\begin{figure}[htb!]
    \centering
    \includegraphics[width=0.95\textwidth]{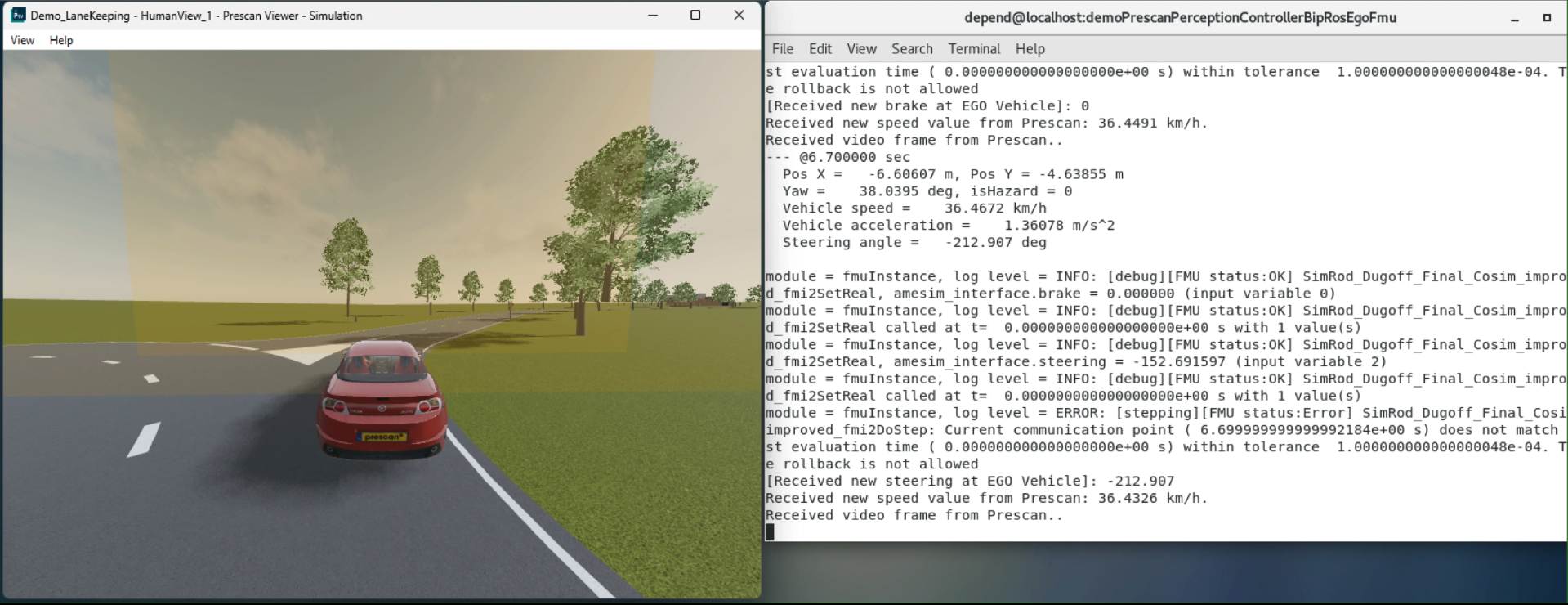}
    \caption{Digital twin simulation of a learning-enabled vehicle  showcasing successful  lane keeping on  a curvy road.}
    \label{fig:rec-3-label}
\end{figure}

In this section, we conduct simulations of the digital twin, highlighting key moments through screenshots captured from a 3D-simulated Prescan video. 
Figure \ref{fig:rec-1-label} shows the successful detection of a traffic sign and the specific $40 km/h$ speed limit. Figure \ref{fig:rec-2-label} shows the successful detection of another traffic sign, this time of and a $60 km/h$ speed limit. Figure \ref{fig:rec-3-label} captures a successful maneuver on a  road with a sharp curvature.

\section{Conclusion}\label{sec:conclusion}

In this paper, we introduced a digital twin prototype of  a learning-enabled self-driving vehicle. The prototype comprises four clients, each representing subsystems/models, collectively contributing to the simulation of real-world driving scenarios. The Amesim client describes the ego vehicle model, the Prescan client constructs a virtual environment, the ROS client guides the ego vehicle through this environment, and the BIP client manages both  the perception module  and the vehicle’s speed control. %The FMU client mirrors the dynamics of the ego vehicle, providing a realistic representation of its behavior. 
The digital twin orchestrator ties all these components together, ensuring efficient data routing and synchronization. We opt for the PAVE360-VSI digital twin platform as it supports both FMI and SystemC-TLM open standards. 
Alternative platforms, such as VICO~\cite{hatledal2021vico}, MAESTRO~\cite{thule2019maestro}, MAESTRO2~\cite{hansen2023co}, DESYRE~\cite{mignogna2013sos}, DIGITBrain~\cite{talasila2021comparison}, and HUBCAP~\cite{macedo2021facilitating}, also offer options for FMI-based co-simulation or SystemC co-simulation of digital twins.
~\\

This DT prototype system highlights the potential of digital twins of learning-enabled autonomous vehicles, demonstrating their ability to  replicate and analyze complex real-world scenarios. A promising future direction  involves the seamless integration of learning-enabled components in the DTs without imposing large computational burdens. Techniques such as neural network abstraction and compression could offer viable solutions, as discussed in works like\cite{mishra2020survey,eleftheriadis2022neural}. Another research direction entails the efficient integration of formal verification methods to facilitate  reasoning about the  conditions under which the DTs satisfy given  safety, security, or performance specifications. Depending on the DT task and application, exhaustive methods \cite{wright2022formally, couto2018injecting, frehse2018toolchain,kekatos2018formal,lisboa2023towards} or non-exhaustive but lightweight methods \cite{temperekidis2022runtime,temperekidis2022towards,dobaj2022towards,havelund2020first} can be considered.

\bibliography{literature}

\begin{thebibliography}{45}
\expandafter\ifx\csname natexlab\endcsname\relax\def\natexlab#1{#1}\fi
\providecommand{\url}[1]{\texttt{#1}}
\providecommand{\href}[2]{#2}
\providecommand{\path}[1]{#1}
\providecommand{\DOIprefix}{doi:}
\providecommand{\ArXivprefix}{arXiv:}
\providecommand{\URLprefix}{URL: }
\providecommand{\Pubmedprefix}{pmid:}
\providecommand{\doi}[1]{\href{http://dx.doi.org/#1}{\path{#1}}}
\providecommand{\Pubmed}[1]{\href{pmid:#1}{\path{#1}}}
\providecommand{\bibinfo}[2]{#2}
\ifx\xfnm\relax \def\xfnm[#1]{\unskip,\space#1}\fi
%Type = Article
\bibitem[{Grieves(2016)}]{grieves2016origins}
\bibinfo{author}{M.~Grieves},
\newblock \bibinfo{title}{Origins of the digital twin concept},
\newblock \bibinfo{journal}{Florida Institute of Technology}
  \bibinfo{volume}{8} (\bibinfo{year}{2016}).
%Type = Article
\bibitem[{Grieves(2019)}]{grieves2019virtually}
\bibinfo{author}{M.~W. Grieves},
\newblock \bibinfo{title}{Virtually intelligent product systems: Digital and
  physical twins},
\newblock \bibinfo{journal}{Complex Systems Engineering: Theory and Practice}
  (\bibinfo{year}{2019}). \URLprefix
  \url{https://api.semanticscholar.org/CorpusID:202478997}.
%Type = Article
\bibitem[{Grieves(2015)}]{Grieves15}
\bibinfo{author}{M.~Grieves},
\newblock \bibinfo{title}{{Digital Twin: Manufacturing Excellence through
  Virtual Factory Replication-A Whitepaper by Dr. Michael Grieves}},
\newblock \bibinfo{journal}{White Paper}  (\bibinfo{year}{2015})
  \bibinfo{pages}{1--7}.
%Type = Article
\bibitem[{Grieves and Vickers(2017)}]{grieves2017digital}
\bibinfo{author}{M.~Grieves}, \bibinfo{author}{J.~Vickers},
\newblock \bibinfo{title}{Digital twin: Mitigating unpredictable, undesirable
  emergent behavior in complex systems},
\newblock \bibinfo{journal}{Transdisciplinary perspectives on complex systems:
  New findings and approaches}  (\bibinfo{year}{2017})
  \bibinfo{pages}{85--113}.
%Type = Incollection
\bibitem[{Grieves(2023)}]{grieves2023digital}
\bibinfo{author}{M.~W. Grieves},
\newblock \bibinfo{title}{Digital twins: Past, present, and future},
\newblock in: \bibinfo{booktitle}{The Digital Twin},
  \bibinfo{publisher}{Springer}, \bibinfo{year}{2023}, pp.
  \bibinfo{pages}{97--121}.
%Type = Article
\bibitem[{Jones et~al.(2020)Jones, Snider, Nassehi, Yon, and
  Hicks}]{jones2020characterising}
\bibinfo{author}{D.~Jones}, \bibinfo{author}{C.~Snider},
  \bibinfo{author}{A.~Nassehi}, \bibinfo{author}{J.~Yon},
  \bibinfo{author}{B.~Hicks},
\newblock \bibinfo{title}{Characterising the digital twin: A systematic
  literature review},
\newblock \bibinfo{journal}{CIRP Journal of Manufacturing Science and
  Technology} \bibinfo{volume}{29} (\bibinfo{year}{2020})
  \bibinfo{pages}{36--52}.
%Type = Article
\bibitem[{Allamaa et~al.(2022)Allamaa, Patrinos, Van~der Auweraer, and
  Son}]{allamaa2022sim2real}
\bibinfo{author}{J.~P. Allamaa}, \bibinfo{author}{P.~Patrinos},
  \bibinfo{author}{H.~Van~der Auweraer}, \bibinfo{author}{T.~D. Son},
\newblock \bibinfo{title}{Sim2real for autonomous vehicle control using
  executable digital twin},
\newblock \bibinfo{journal}{IFAC-PapersOnLine} \bibinfo{volume}{55}
  (\bibinfo{year}{2022}) \bibinfo{pages}{385--391}.
%Type = Article
\bibitem[{Piromalis and Kantaros(2022)}]{piromalis2022digital}
\bibinfo{author}{D.~Piromalis}, \bibinfo{author}{A.~Kantaros},
\newblock \bibinfo{title}{Digital twins in the automotive industry: The road
  toward physical-digital convergence},
\newblock \bibinfo{journal}{Applied System Innovation} \bibinfo{volume}{5}
  (\bibinfo{year}{2022}) \bibinfo{pages}{65}.
%Type = Article
\bibitem[{Esen and Liao(2023)}]{esen2023simulation}
\bibinfo{author}{H.~Esen}, \bibinfo{author}{B.~H.-C. Liao},
\newblock \bibinfo{title}{Simulation-based safety assurance for an avp system
  incorporating learning-enabled components},
\newblock \bibinfo{journal}{arXiv preprint arXiv:2311.03362}
  (\bibinfo{year}{2023}).
%Type = Inproceedings
\bibitem[{T{\o}ttrup et~al.(2022)T{\o}ttrup, Hu, Kramer, Macedo, and
  Esterle}]{tottrup2022using}
\bibinfo{author}{M.~F. T{\o}ttrup}, \bibinfo{author}{E.~C. Hu},
  \bibinfo{author}{B.~A. Kramer}, \bibinfo{author}{H.~D. Macedo},
  \bibinfo{author}{L.~Esterle},
\newblock \bibinfo{title}{{Using INTO-CPS Tools in the Development of a Digital
  Twin for the F1TENTH Race Car}},
\newblock in: \bibinfo{booktitle}{International Conference on Software
  Engineering and Formal Methods}, \bibinfo{organization}{Springer},
  \bibinfo{year}{2022}, pp. \bibinfo{pages}{200--209}.
%Type = Article
\bibitem[{Chaudhuri et~al.(2023)Chaudhuri, Pash, Hormuth, Lorenzo, Kapteyn, Wu,
  Lima, Yankeelov, Willcox et~al.}]{chaudhuri2023predictive}
\bibinfo{author}{A.~Chaudhuri}, \bibinfo{author}{G.~Pash},
  \bibinfo{author}{D.~A. Hormuth}, \bibinfo{author}{G.~Lorenzo},
  \bibinfo{author}{M.~Kapteyn}, \bibinfo{author}{C.~Wu}, \bibinfo{author}{E.~A.
  Lima}, \bibinfo{author}{T.~E. Yankeelov}, \bibinfo{author}{K.~Willcox},
  et~al.,
\newblock \bibinfo{title}{Predictive digital twin for optimizing
  patient-specific radiotherapy regimens under uncertainty in high-grade
  gliomas},
\newblock \bibinfo{journal}{Frontiers in Artificial Intelligence}
  \bibinfo{volume}{6} (\bibinfo{year}{2023}).
%Type = Article
\bibitem[{Hartmann and Van~der Auweraer(2022)}]{hartmann2022executable}
\bibinfo{author}{D.~Hartmann}, \bibinfo{author}{H.~Van~der Auweraer},
\newblock \bibinfo{title}{The executable digital twin: merging the digital and
  the physics worlds},
\newblock \bibinfo{journal}{arXiv preprint arXiv:2210.17402}
  (\bibinfo{year}{2022}).
%Type = Article
\bibitem[{Torzoni et~al.(2024)Torzoni, Tezzele, Mariani, Manzoni, and
  Willcox}]{torzoni2024digital}
\bibinfo{author}{M.~Torzoni}, \bibinfo{author}{M.~Tezzele},
  \bibinfo{author}{S.~Mariani}, \bibinfo{author}{A.~Manzoni},
  \bibinfo{author}{K.~E. Willcox},
\newblock \bibinfo{title}{A digital twin framework for civil engineering
  structures},
\newblock \bibinfo{journal}{Computer Methods in Applied Mechanics and
  Engineering} \bibinfo{volume}{418} (\bibinfo{year}{2024})
  \bibinfo{pages}{116584}.
%Type = Article
\bibitem[{Frasheri et~al.(2023)Frasheri, Ejersbo, Thule, Gomes, Kvistgaard,
  Larsen, and Esterle}]{frasheri2023addressing}
\bibinfo{author}{M.~Frasheri}, \bibinfo{author}{H.~Ejersbo},
  \bibinfo{author}{C.~Thule}, \bibinfo{author}{C.~Gomes},
  \bibinfo{author}{J.~L. Kvistgaard}, \bibinfo{author}{P.~G. Larsen},
  \bibinfo{author}{L.~Esterle},
\newblock \bibinfo{title}{Addressing time discrepancy between digital and
  physical twins},
\newblock \bibinfo{journal}{Robotics and Autonomous Systems}
  \bibinfo{volume}{161} (\bibinfo{year}{2023}) \bibinfo{pages}{104347}.
%Type = Article
\bibitem[{Deakin et~al.(2023)Deakin, Vanin, Fan, and
  Van~Hertem}]{deakin2023smart}
\bibinfo{author}{M.~Deakin}, \bibinfo{author}{M.~Vanin},
  \bibinfo{author}{Z.~Fan}, \bibinfo{author}{D.~Van~Hertem},
\newblock \bibinfo{title}{Smart energy network digital twins: Findings from a
  uk-based demonstrator project},
\newblock \bibinfo{journal}{arXiv preprint arXiv:2311.11997}
  (\bibinfo{year}{2023}).
%Type = Inproceedings
\bibitem[{Bensalem et~al.(2024)Bensalem, Katsaros, Ni{\v{c}}kovi{\'{c}}, Liao,
  Nolasco, AbdElSalam, Beyene, Cano, Delacourt, Esen, Forrai, He, Huang,
  Kekatos, K{\"o}nighofer, Paulitsch, Peled, Ponchant, Sorokin, Tong, and
  Wu}]{10.1007/978-3-031-46002-9_15}
\bibinfo{author}{S.~Bensalem}, \bibinfo{author}{P.~Katsaros},
  \bibinfo{author}{D.~Ni{\v{c}}kovi{\'{c}}}, \bibinfo{author}{B.~H.-C. Liao},
  \bibinfo{author}{R.~R. Nolasco}, \bibinfo{author}{M.~AbdElSalam},
  \bibinfo{author}{T.~Beyene}, \bibinfo{author}{F.~Cano},
  \bibinfo{author}{A.~Delacourt}, \bibinfo{author}{H.~Esen},
  \bibinfo{author}{A.~Forrai}, \bibinfo{author}{W.~He},
  \bibinfo{author}{X.~Huang}, \bibinfo{author}{N.~Kekatos},
  \bibinfo{author}{B.~K{\"o}nighofer}, \bibinfo{author}{M.~Paulitsch},
  \bibinfo{author}{D.~Peled}, \bibinfo{author}{M.~Ponchant},
  \bibinfo{author}{L.~Sorokin}, \bibinfo{author}{S.~Tong},
  \bibinfo{author}{C.~Wu},
\newblock \bibinfo{title}{Continuous engineering for trustworthy
  learning-enabled autonomous systems},
\newblock in: \bibinfo{editor}{B.~Steffen} (Ed.), \bibinfo{booktitle}{Bridging
  the Gap Between AI and Reality}, \bibinfo{publisher}{Springer Nature
  Switzerland}, \bibinfo{address}{Cham}, \bibinfo{year}{2024}, pp.
  \bibinfo{pages}{256--278}.
%Type = Article
\bibitem[{Gomes et~al.(2017)Gomes, Thule, Broman, Larsen, and
  Vangheluwe}]{gomes2017co}
\bibinfo{author}{C.~Gomes}, \bibinfo{author}{C.~Thule},
  \bibinfo{author}{D.~Broman}, \bibinfo{author}{P.~G. Larsen},
  \bibinfo{author}{H.~Vangheluwe},
\newblock \bibinfo{title}{Co-simulation: State of the art},
\newblock \bibinfo{journal}{arXiv preprint arXiv:1702.00686}
  (\bibinfo{year}{2017}).
%Type = Article
\bibitem[{Gomes et~al.(2018)Gomes, Thule, Larsen, Denil, and
  Vangheluwe}]{gomes2018co}
\bibinfo{author}{C.~Gomes}, \bibinfo{author}{C.~Thule}, \bibinfo{author}{P.~G.
  Larsen}, \bibinfo{author}{J.~Denil}, \bibinfo{author}{H.~Vangheluwe},
\newblock \bibinfo{title}{Co-simulation of continuous systems: a tutorial},
\newblock \bibinfo{journal}{arXiv preprint arXiv:1809.08463}
  (\bibinfo{year}{2018}).
%Type = Article
\bibitem[{Talasila et~al.(2023)Talasila, Gomes, Mikkelsen, Arboleda, Kamburjan,
  and Larsen}]{talasila2023digital}
\bibinfo{author}{P.~Talasila}, \bibinfo{author}{C.~Gomes},
  \bibinfo{author}{P.~H. Mikkelsen}, \bibinfo{author}{S.~G. Arboleda},
  \bibinfo{author}{E.~Kamburjan}, \bibinfo{author}{P.~G. Larsen},
\newblock \bibinfo{title}{{Digital Twin as a Service (DTaaS): A Platform for
  Digital Twin Developers and Users}},
\newblock \bibinfo{journal}{arXiv preprint arXiv:2305.07244}
  (\bibinfo{year}{2023}).
%Type = Inproceedings
\bibitem[{Blockwitz et~al.(2012)Blockwitz, Otter, Akesson, Arnold, Clauss,
  Elmqvist, Friedrich, Junghanns, Mauss, Neumerkel
  et~al.}]{blockwitz2012functional}
\bibinfo{author}{T.~Blockwitz}, \bibinfo{author}{M.~Otter},
  \bibinfo{author}{J.~Akesson}, \bibinfo{author}{M.~Arnold},
  \bibinfo{author}{C.~Clauss}, \bibinfo{author}{H.~Elmqvist},
  \bibinfo{author}{M.~Friedrich}, \bibinfo{author}{A.~Junghanns},
  \bibinfo{author}{J.~Mauss}, \bibinfo{author}{D.~Neumerkel}, et~al.,
\newblock \bibinfo{title}{Functional mockup interface 2.0: The standard for
  tool independent exchange of simulation models},
\newblock in: \bibinfo{booktitle}{Proceedings of the 9th International Modelica
  Conference}, \bibinfo{publisher}{Munich, Germany}, \bibinfo{year}{2012}, pp.
  \bibinfo{pages}{173--184}.
%Type = Inproceedings
\bibitem[{Junghanns et~al.(2021)Junghanns, Gomes, Schulze, Schuch, Pierre,
  Blaesken, Zacharias, Pillekeit, Wernersson, Sommer
  et~al.}]{junghanns2021functional}
\bibinfo{author}{A.~Junghanns}, \bibinfo{author}{C.~Gomes},
  \bibinfo{author}{C.~Schulze}, \bibinfo{author}{K.~Schuch},
  \bibinfo{author}{R.~Pierre}, \bibinfo{author}{M.~Blaesken},
  \bibinfo{author}{I.~Zacharias}, \bibinfo{author}{A.~Pillekeit},
  \bibinfo{author}{K.~Wernersson}, \bibinfo{author}{T.~Sommer}, et~al.,
\newblock \bibinfo{title}{The functional mock-up interface 3.0-new features
  enabling new applications},
\newblock in: \bibinfo{booktitle}{Modelica conferences}, \bibinfo{year}{2021},
  pp. \bibinfo{pages}{17--26}.
%Type = Article
\bibitem[{Frank(2005)}]{frank2005transaction}
\bibinfo{author}{G.~Frank},
\newblock \bibinfo{title}{{Transaction-Level Modeling with SystemC: TLM
  Concepts and Applications for Embedded Systems}},
\newblock \bibinfo{journal}{Springer}  (\bibinfo{year}{2005}).
%Type = Inproceedings
\bibitem[{Basu et~al.(2011)Basu, Bensalem, Bozga, Bourgos, and
  Sifakis}]{basu2011rigorous}
\bibinfo{author}{A.~Basu}, \bibinfo{author}{S.~Bensalem},
  \bibinfo{author}{M.~Bozga}, \bibinfo{author}{P.~Bourgos},
  \bibinfo{author}{J.~Sifakis},
\newblock \bibinfo{title}{{Rigorous system design: the BIP approach}},
\newblock in: \bibinfo{booktitle}{International doctoral workshop on
  mathematical and engineering methods in computer science},
  \bibinfo{organization}{Springer}, \bibinfo{year}{2011}, pp.
  \bibinfo{pages}{1--19}.
%Type = Misc
\bibitem[{{Siemens EDA}(2023)}]{veloce}
\bibinfo{author}{{Siemens EDA}},
  \bibinfo{title}{{Veloce\textsuperscript{\textregistered}}},
  \bibinfo{howpublished}{Available at:
  \url{https://eda.sw.siemens.com/en-US/ic/veloce/}}, \bibinfo{year}{2023}.
%Type = Article
\bibitem[{AbdElSalam et~al.(2019)AbdElSalam, Khalil, Stickley, Salem, and
  Loye}]{abdelsalam2019verification}
\bibinfo{author}{M.~AbdElSalam}, \bibinfo{author}{K.~Khalil},
  \bibinfo{author}{J.~Stickley}, \bibinfo{author}{A.~Salem},
  \bibinfo{author}{B.~Loye},
\newblock \bibinfo{title}{Verification of advanced driver assistance systems
  and autonomous vehicles with hardware emulation-in-the-loop a case study with
  multiple ecus},
\newblock \bibinfo{journal}{IJAE}  (\bibinfo{year}{2019}).
%Type = Misc
\bibitem[{{Adam Erickson, John Stickley}(2023)}]{uvm}
\bibinfo{author}{{Adam Erickson, John Stickley}}, \bibinfo{title}{{UVM-Connect
  primer}}, \bibinfo{howpublished}{Available at:
  \url{https://verificationacademy.com/courses/uvm-connect}},
  \bibinfo{year}{2023}.
%Type = Inproceedings
\bibitem[{Temperekidis et~al.(2022{\natexlab{a}})Temperekidis, Kekatos,
  Katsaros, He, Bensalem, AbdElSabour, AbdElSalam, and
  Salem}]{temperekidis2022towards}
\bibinfo{author}{A.~Temperekidis}, \bibinfo{author}{N.~Kekatos},
  \bibinfo{author}{P.~Katsaros}, \bibinfo{author}{W.~He},
  \bibinfo{author}{S.~Bensalem}, \bibinfo{author}{H.~AbdElSabour},
  \bibinfo{author}{M.~AbdElSalam}, \bibinfo{author}{A.~Salem},
\newblock \bibinfo{title}{Towards a digital twin architecture with formal
  analysis capabilities for learning-enabled autonomous systems},
\newblock in: \bibinfo{booktitle}{International Conference on Modelling and
  Simulation for Autonomous Systems}, \bibinfo{organization}{Springer},
  \bibinfo{year}{2022}{\natexlab{a}}, pp. \bibinfo{pages}{163--181}.
%Type = Inproceedings
\bibitem[{Temperekidis et~al.(2022{\natexlab{b}})Temperekidis, Kekatos, and
  Katsaros}]{temperekidis2022runtime}
\bibinfo{author}{A.~Temperekidis}, \bibinfo{author}{N.~Kekatos},
  \bibinfo{author}{P.~Katsaros},
\newblock \bibinfo{title}{{Runtime verification for FMI-based co-simulation}},
\newblock in: \bibinfo{booktitle}{International Conference on Runtime
  Verification}, \bibinfo{organization}{Springer},
  \bibinfo{year}{2022}{\natexlab{b}}, pp. \bibinfo{pages}{304--313}.
%Type = Misc
\bibitem[{{Siemens PLM Software}(2023)}]{amesim}
\bibinfo{author}{{Siemens PLM Software}}, \bibinfo{title}{{Simcenter Amesim}},
  \bibinfo{howpublished}{Available at:
  \url{https://www.plm.automation.siemens.com/en/products/lms/imagine-lab/amesim/}},
  \bibinfo{year}{2023}.
%Type = Article
\bibitem[{Ge et~al.(2021)Ge, Liu, Wang, Li, and Sun}]{ge2021yolox}
\bibinfo{author}{Z.~Ge}, \bibinfo{author}{S.~Liu}, \bibinfo{author}{F.~Wang},
  \bibinfo{author}{Z.~Li}, \bibinfo{author}{J.~Sun},
\newblock \bibinfo{title}{{Yolox: Exceeding yolo series in 2021}},
\newblock \bibinfo{journal}{arXiv preprint arXiv:2107.08430}
  (\bibinfo{year}{2021}).
%Type = Article
\bibitem[{Hatledal et~al.(2021)Hatledal, Chu, Styve, and
  Zhang}]{hatledal2021vico}
\bibinfo{author}{L.~I. Hatledal}, \bibinfo{author}{Y.~Chu},
  \bibinfo{author}{A.~Styve}, \bibinfo{author}{H.~Zhang},
\newblock \bibinfo{title}{{Vico: An entity-component-system based co-simulation
  framework}},
\newblock \bibinfo{journal}{Simulation Modelling Practice and Theory}
  \bibinfo{volume}{108} (\bibinfo{year}{2021}) \bibinfo{pages}{102243}.
%Type = Article
\bibitem[{Thule et~al.(2019)Thule, Lausdahl, Gomes, Meisl, and
  Larsen}]{thule2019maestro}
\bibinfo{author}{C.~Thule}, \bibinfo{author}{K.~Lausdahl},
  \bibinfo{author}{C.~Gomes}, \bibinfo{author}{G.~Meisl},
  \bibinfo{author}{P.~G. Larsen},
\newblock \bibinfo{title}{Maestro: the into-cps co-simulation framework},
\newblock \bibinfo{journal}{Simulation Modelling Practice and Theory}
  \bibinfo{volume}{92} (\bibinfo{year}{2019}) \bibinfo{pages}{45--61}.
%Type = Article
\bibitem[{Hansen et~al.(2023)Hansen, Thule, Gomes, Lausdahl, Madsen, Abbiati,
  and Larsen}]{hansen2023co}
\bibinfo{author}{S.~T. Hansen}, \bibinfo{author}{C.~Thule},
  \bibinfo{author}{C.~Gomes}, \bibinfo{author}{K.~G. Lausdahl},
  \bibinfo{author}{F.~P. Madsen}, \bibinfo{author}{G.~Abbiati},
  \bibinfo{author}{P.~G. Larsen},
\newblock \bibinfo{title}{{Co-simulation at different levels of expertise with
  Maestro2}},
\newblock \bibinfo{journal}{Journal of Systems and Software}
  (\bibinfo{year}{2023}) \bibinfo{pages}{111905}.
%Type = Article
\bibitem[{Mignogna et~al.(2013)Mignogna, Mangeruca, Boyer, Legay, and
  Arnold}]{mignogna2013sos}
\bibinfo{author}{A.~Mignogna}, \bibinfo{author}{L.~Mangeruca},
  \bibinfo{author}{B.~Boyer}, \bibinfo{author}{A.~Legay},
  \bibinfo{author}{A.~Arnold},
\newblock \bibinfo{title}{Sos contract verification using statistical model
  checking},
\newblock \bibinfo{journal}{arXiv preprint arXiv:1311.3632}
  (\bibinfo{year}{2013}).
%Type = Inproceedings
\bibitem[{Talasila et~al.(2021)Talasila, Cr{\u{a}}ciunean, Bogdan-Constantin,
  Larsen, Zamfirescu, and Scovill}]{talasila2021comparison}
\bibinfo{author}{P.~Talasila}, \bibinfo{author}{D.-C. Cr{\u{a}}ciunean},
  \bibinfo{author}{P.~Bogdan-Constantin}, \bibinfo{author}{P.~G. Larsen},
  \bibinfo{author}{C.~Zamfirescu}, \bibinfo{author}{A.~Scovill},
\newblock \bibinfo{title}{Comparison between the hubcap and digitbrain
  platforms for model-based design and evaluation of digital twins},
\newblock in: \bibinfo{booktitle}{International Conference on Software
  Engineering and Formal Methods}, \bibinfo{organization}{Springer},
  \bibinfo{year}{2021}, pp. \bibinfo{pages}{238--244}.
%Type = Article
\bibitem[{Macedo et~al.(2021)Macedo, Sassanelli, Larsen, and
  Terzi}]{macedo2021facilitating}
\bibinfo{author}{H.~D. Macedo}, \bibinfo{author}{C.~Sassanelli},
  \bibinfo{author}{P.~G. Larsen}, \bibinfo{author}{S.~Terzi},
\newblock \bibinfo{title}{Facilitating model-based design of
  cyber-manufacturing systems},
\newblock \bibinfo{journal}{Procedia CIRP} \bibinfo{volume}{104}
  (\bibinfo{year}{2021}) \bibinfo{pages}{1936--1941}.
%Type = Article
\bibitem[{Mishra et~al.(2020)Mishra, Gupta, and Dutta}]{mishra2020survey}
\bibinfo{author}{R.~Mishra}, \bibinfo{author}{H.~P. Gupta},
  \bibinfo{author}{T.~Dutta},
\newblock \bibinfo{title}{A survey on deep neural network compression:
  Challenges, overview, and solutions},
\newblock \bibinfo{journal}{arXiv preprint arXiv:2010.03954}
  (\bibinfo{year}{2020}).
%Type = Inproceedings
\bibitem[{Eleftheriadis et~al.(2022)Eleftheriadis, Kekatos, Katsaros, and
  Tripakis}]{eleftheriadis2022neural}
\bibinfo{author}{C.~Eleftheriadis}, \bibinfo{author}{N.~Kekatos},
  \bibinfo{author}{P.~Katsaros}, \bibinfo{author}{S.~Tripakis},
\newblock \bibinfo{title}{On neural network equivalence checking using smt
  solvers},
\newblock in: \bibinfo{booktitle}{International Conference on Formal Modeling
  and Analysis of Timed Systems}, \bibinfo{organization}{Springer},
  \bibinfo{year}{2022}, pp. \bibinfo{pages}{237--257}.
%Type = Inproceedings
\bibitem[{Wright et~al.(2022)Wright, Gomes, and Woodcock}]{wright2022formally}
\bibinfo{author}{T.~Wright}, \bibinfo{author}{C.~Gomes},
  \bibinfo{author}{J.~Woodcock},
\newblock \bibinfo{title}{Formally verified self-adaptation of an incubator
  digital twin},
\newblock in: \bibinfo{booktitle}{International Symposium on Leveraging
  Applications of Formal Methods}, \bibinfo{organization}{Springer},
  \bibinfo{year}{2022}, pp. \bibinfo{pages}{89--109}.
%Type = Inproceedings
\bibitem[{Couto et~al.(2018)Couto, Basagiannis, Ridouane, Mady, Hasanagic, and
  Larsen}]{couto2018injecting}
\bibinfo{author}{L.~D. Couto}, \bibinfo{author}{S.~Basagiannis},
  \bibinfo{author}{E.~H. Ridouane}, \bibinfo{author}{A.~E.-D. Mady},
  \bibinfo{author}{M.~Hasanagic}, \bibinfo{author}{P.~G. Larsen},
\newblock \bibinfo{title}{Injecting formal verification in fmi-based
  co-simulations of cyber-physical systems},
\newblock in: \bibinfo{booktitle}{Software Engineering and Formal Methods},
  \bibinfo{organization}{Springer}, \bibinfo{year}{2018}, pp.
  \bibinfo{pages}{284--299}.
%Type = Inproceedings
\bibitem[{Frehse et~al.(2018)Frehse, Kekatos, Nickovic, Oehlerking, Schuler,
  Walsch, and Woehrle}]{frehse2018toolchain}
\bibinfo{author}{G.~Frehse}, \bibinfo{author}{N.~Kekatos},
  \bibinfo{author}{D.~Nickovic}, \bibinfo{author}{J.~Oehlerking},
  \bibinfo{author}{S.~Schuler}, \bibinfo{author}{A.~Walsch},
  \bibinfo{author}{M.~Woehrle},
\newblock \bibinfo{title}{A toolchain for verifying safety properties of hybrid
  automata via pattern templates},
\newblock in: \bibinfo{booktitle}{2018 Annual American Control Conference
  (ACC)}, \bibinfo{organization}{IEEE}, \bibinfo{year}{2018}, pp.
  \bibinfo{pages}{2384--2391}.
%Type = Phdthesis
\bibitem[{Kekatos(2018)}]{kekatos2018formal}
\bibinfo{author}{N.~Kekatos}, \bibinfo{title}{Formal verification of
  cyber-physical systems in the industrial model-based design process}, Ph.D.
  thesis, Universit{\'e} Grenoble Alpes, \bibinfo{year}{2018}.
%Type = Article
\bibitem[{Lisboa~Malaquias et~al.(2023)Lisboa~Malaquias, Giantamidis,
  Basagiannis, Fulvio~Rollini, and Amundson}]{lisboa2023towards}
\bibinfo{author}{F.~Lisboa~Malaquias}, \bibinfo{author}{G.~Giantamidis},
  \bibinfo{author}{S.~Basagiannis}, \bibinfo{author}{S.~Fulvio~Rollini},
  \bibinfo{author}{I.~Amundson},
\newblock \bibinfo{title}{Towards a methodology to design provably secure
  cyber-physical systems},
\newblock \bibinfo{journal}{ACM SIGAda Ada Letters} \bibinfo{volume}{43}
  (\bibinfo{year}{2023}) \bibinfo{pages}{94--99}.
%Type = Inproceedings
\bibitem[{Dobaj et~al.(2022)Dobaj, Riel, Krug, Seidl, Macher, and
  Egretzberger}]{dobaj2022towards}
\bibinfo{author}{J.~Dobaj}, \bibinfo{author}{A.~Riel},
  \bibinfo{author}{T.~Krug}, \bibinfo{author}{M.~Seidl},
  \bibinfo{author}{G.~Macher}, \bibinfo{author}{M.~Egretzberger},
\newblock \bibinfo{title}{Towards digital twin-enabled devops for cps providing
  architecture-based service adaptation \& verification at runtime},
\newblock in: \bibinfo{booktitle}{Proceedings of the 17th Symposium on Software
  Engineering for Adaptive and Self-Managing Systems}, \bibinfo{year}{2022},
  pp. \bibinfo{pages}{132--143}.
%Type = Article
\bibitem[{Havelund et~al.(2020)Havelund, Peled, and Ulus}]{havelund2020first}
\bibinfo{author}{K.~Havelund}, \bibinfo{author}{D.~Peled},
  \bibinfo{author}{D.~Ulus},
\newblock \bibinfo{title}{First-order temporal logic monitoring with bdds},
\newblock \bibinfo{journal}{Formal Methods in System Design}
  \bibinfo{volume}{56} (\bibinfo{year}{2020}) \bibinfo{pages}{1--21}.

\end{thebibliography}

\end{document}